# Computational Interpretations of the Gricean Maxims in the Generation of Referring Expressions[*]

Robert Dale[†]
Microsoft Institute
for Advanced Software Technology
North Ryde
Sydney NSW 2113
Australia

Ehud Reiter[‡]
CoGenTex Inc
840 Hanshaw Rd
Ithaca
NY 14850
USA

September 27, 1994

**Abstract**

We examine the problem of generating definite noun phrases that are appropriate referring expressions; that is, noun phrases that (1) successfully identify the intended referent to the hearer whilst (2) not conveying to her any false conversational implicatures (Grice, 1975). We review several possible computational interpretations of the conversational implicature maxims, with different computational costs, and argue that the simplest may be the best, because it seems to be closest to what human speakers do. We describe our recommended algorithm in detail, along with a specification of the resources a host system must provide in order to make use of the algorithm, and an implementation used in the natural language generation component of the IDAS system.

## 1 Introduction

One of the most ubiquitous tasks in natural language generation is the generation of referring expressions: phrases that identify particular domain entities to the human recipient of the generation system's output. In this paper, we examine the task of generating one particular kind of referring expression, focusing on computational, algorithmic, and pragmatic issues that would be more difficult to investigate in a broader study. Our hope is that the conclusions we draw from this 'micro-study', in particular that computationally simple interpretations of the Gricean maxims of conversational implicature (Grice, 1975) should be used, will also prove applicable to other referring expression generation tasks.

---

The particular referring expressions which we study in this paper have the following characteristics.

1. They are linguistically realized as definite noun phrases (for example, *the red cup*), rather than pronouns, *one*-anaphora, and other alternative linguistic mechanisms used for reference.
2. They refer to physical objects (for example, dogs and tables), rather than abstract entities such as fields of mathematics.
3. They are solely intended to identify the target object to the hearer, and are not intended to satisfy any other communicative goal.

*The black dog*, *the small skinny screw*, and *the upside-down cup* are examples of the kind of referring expressions which we included in our study. *One*-anaphoric expressions (such as *the red one*), nominalisations (such as *the death of Caesar*), and noun phrases containing material that performs some communicative function other than referent identification (as in *a dangerous and hungry shark*) are examples of referring expressions which we did not examine in this study; we have, however, investigated issues involved in generating such referring expressions in other work (Dale, 1992; Oberlander and Dale, 1991; and Reiter, 1990c respectively).

There are many criteria that an algorithm which generates referring expressions should satisfy; we are particularly concerned here with the following:

1. The algorithm should generate referring expressions which satisfy the referential communicative goal: after hearing or seeing the referring expression, the human hearer or reader should be able to identify the target object.
2. The algorithm should generate referring expressions which do not lead the human hearer or reader to make false conversational implicatures in the sense of Grice (1975).
3. The algorithm, if it is to be of practical use, should be computationally efficient.

Previous work has tended to focus only on the first of these criteria. Our goal was to take into account the second and third criteria as well, and in particular to investigate and evaluate different possible computational interpretations of the Gricean maxims of conversational implicature. This evaluation was based on (a) the computational cost of obeying each interpretation, and (b) how closely each interpretation approximates the behaviour of human speakers. We were also interested in establishing which aspects of the process of referring expression generation can be characterised by a general purpose algorithm, and which aspects are best left to domain-dependent specification.

The structure of the paper is as follows. In Section 2, we survey previous work in the generation of referring expressions, and elaborate on the issues of concern in the present work. In Section 3, we examine four different computational interpretations of Grice's conversational maxims in the context of the generation of referring expressions; we discuss both the computational complexity (expense) of these interpretations, and how they compare to what is known about how human speakers generate referring expressions. In Section 4, we discuss in more detail the interpretation we find most appealing, because it is both fastest and (as far as we can tell) closest to what human speakers do; we present this algorithm in a domain-independent way, together with a description of the domain knowledge resources that the algorithm needs. In Section 5 we present some concluding comments and discuss briefly a number of areas in which more research needs to be done.

## 2 The Issues

### 2.1 Previous Work

Perhaps the best-known work in generating referring expressions has been done by Appelt and Kronfeld (Appelt, 1985a; Appelt, 1985b; Appelt, 1985c; Kronfeld, 1986; Appelt and Kronfeld, 1987). Compared to our work, Appelt and Kronfeld allowed much broader communicative goals, but on the other hand paid relatively little attention to conversational implicature and (especially) computational cost. Some examples of referring expressions they studied which had goals beyond simple identification are as follows.

(1)     *That scoundrel* is the one who betrayed us.

(2)     I met *an old friend* yesterday.

In (1), describing the intended referent as a *scoundrel* has the effect of informing the hearer that the speaker has a very low opinion of the person in question; it is not intended to help the hearer identify a particular person from a set of people. In (2), the speaker informs the hearer that the person she met was an old friend, but does not intend for the hearer to be able to determine the exact identity of the person.

Appelt's thesis work (Appelt, 1985a; Appelt, 1985b) in fact allowed 'referring expressions' to satisfy any communicative goal that could be stated in the underlying logical framework. Appelt constructed such referring expressions using standard AI planning and theorem-proving techniques, with very little concern being paid to computational efficiency issues. He also said relatively little about conversational implicature, except to state that generations systems should attempt to generate short referring expressions with a heuristic algorithm.

Goodman (1986) has looked at cases where inappropriate referring expressions are generated by human speakers, and proposed a 'Find What I Mean' model for determining what object the speaker intended to refer to; this work is more relevant to understanding referring expressions than to generating them. Goodman was primarily interested in handling situations where a referring expression was not a distinguishing description in the sense explained below; although he briefly mentions conversational implicature and overly-specific descriptions, he does not seem to have looked at these cases in as much detail.

The previous work of the present authors on generating definite NPs that identify physical objects is discussed in Section 3.1. We have also worked on generating other types of referring expressions. Dale (1992) has looked at the generation of *one*-anaphoric referring expressions; and Dale and Haddock (1991) have examined the generation of referring expressions that include relations, as opposed to simple predicative properties. Oberlander and Dale (1991) have looked at the problem of referring to eventualities. Reiter (1990c) has looked at the problem of generating NPs that are intended to inform the hearer that an entity has certain properties, as opposed to simply identifying the entity.

## 2.2 Satisfying the Referential Communicative Goal

In this paper, we are only concerned with referring expressions whose sole purpose is to identify an entity to the hearer.[1] We follow Dale and Haddock (1991) in assuming that a referring expression satisfies the referential communicative goal if it is a **distinguishing description**; that is, if it is an accurate description of the entity being referred to, but not of any other object in the current **context set**. We define the context set to be the set of entities that the hearer is currently assumed to be attending to; this is similar to the set of entities in the focus spaces of the discourse focus stack in Grosz and Sidner's theory of discourse structure (Grosz & Sidner, 1986). We also define the **contrast set** to be all elements of the context set except the intended referent; this has also been referred to as the set of **potential distractors** (McDonald, 1981).

Under this model, each property expressed in a referring expression can be regarded as having the function of 'ruling out' members of the contrast set. Suppose a speaker wants to identify a small black dog in a situation where the contrast set consists of a large white dog and a small black cat. She might choose the adjective *black* in order to rule out the white dog, and the head noun *dog* in order to rule out the cat; this would result in the generation of the referring expression *the black dog*, which matches the intended referent but no other object in the current context. *The small dog* would also be a successful referring expression in this context, under the distinguishing description model.

More formally, we assume that each entity in the domain is characterised by means of a set of attribute–value pairs. We will sometimes to refer to an attribute–value pair as a **property**. We assume that the semantic content of a referring expression can also be represented as a set of attribute–value pairs. We will use the notation ⟨Attribute, Value⟩ for attribute–value pairs; for example, ⟨colour, red⟩ indicates the attribute of *colour* with the value *red*. The semantic content of head nouns will be represented as the value of the special attribute type: for example, ⟨type, dog⟩.[2]

Let $r$ be the intended referent, and $C$ be the contrast set; then, a set $L$ of attribute–value

---

[1] We only use what Kronfeld (1986) calls the modal aspect of Donellan's distinction (Donnellan, 1966) between 'referential' and 'attributive' descriptions; that is, we are examining descriptive NPs that are *intended* to identify an object to the hearer.

[2] Any work in natural language generation is faced with the problem of deciding what kind of input representation to start from; it is this that determines in large part how hard or how easy the generation task will be. We assume that each attribute–value pair—that is, each semantic constituent—can be realized as a syntactic constituent; for example, as an adjective such as *red* or as a prepositional phrase such as *with black eyes*. In cases where one lexical item can convey several properties, we assume that this lexical item can be represented by a single attribute–value pair in the semantics, which may require creating a special attribute for this purpose. Thus, for example, we assume that the semantic content of *the bachelor lawyer* is represented as {⟨type, lawyer⟩, ⟨marital-status, bachelor⟩}, rather than by means of more primitive elements such as {⟨type, lawyer⟩, ⟨sex, male⟩, ⟨age, adult⟩, ⟨married, false⟩}. We leave the question of how one might generate from these more primitive representational elements for future work. This is ultimately rather unsatisfactory, of course, because there will also be more complex structures we might want to compose out of the elements we happen to have chosen here as primitives; but in the absence of any consensus on what the 'right' input representation is—whatever that means—we take the view that a level of representation that is 'just below the surface' is more appropriate than one whose elements are some set of semantic primitives based on intuition.

pairs will represent a distinguishing description if the two conditions in (3) hold:

(3)  C1:  Every attribute–value pair in $L$ applies to $r$: that is, every element of $L$ specifies an attribute–value that $r$ possesses.

  C2:  For every member $c$ of $C$, there is at least one element $l$ of $L$ that does not apply to $c$: that is, there is an $l$ in $L$ that specifies an attribute–value that $c$ does not possess. $l$ is said to **rule out** $c$.

For example, suppose the task is to create a referring expression for Object1 in a context that also includes Object2 and Object3, where these objects possess the following properties:

- Object1: ⟨type, dog⟩, ⟨size, small⟩, ⟨colour, black⟩
- Object2: ⟨type, dog⟩, ⟨size, large⟩, ⟨colour, white⟩
- Object3: ⟨type, cat⟩, ⟨size, small⟩, ⟨colour, black⟩

In this situation, $r =$ Object1 and $C = \{$Object2, Object3$\}$. The content of one possible distinguishing description is then $\{⟨$type, dog$⟩, ⟨$colour, black$⟩\}$, which might be realised as *the black dog*: Object1 possesses these properties, but Object2 and Object3 do not (Object2 does not have the property ⟨colour, black⟩, while Object3 is ruled out by ⟨type, dog⟩).

With this definition, finding a distinguishing description is essentially equivalent to solving a **set cover** problem. Let $P$ be the set of properties that $r$ possesses, and let $RulesOut(p)$ be the subset of $C$ that is ruled out by each $p$ in $P$. If $RulesOut(p)$ is not the empty set, then we say that $p$ has some **discriminatory power**.[3] Then, $L$ will be a distinguishing description if the union of $RulesOut(l)$, for every $l$ in $L$, is equal to $C$ (that is, if $L$ specifies a set of $RulesOut$ sets that together **cover** all of $C$).

The identification of the problem of finding a distinguishing description with the problem of finding a set cover is a useful one, because it enables us to use the algorithms and complexity results derived by theoretical computer scientists for set cover problems. For example, researchers in that area have proven that finding the minimal size set cover is NP-Hard (Garey & Johnson, 1979), which tells us that finding the smallest possible referring expression is an NP-Hard task and thus probably computationally impractical. Algorithm researchers have also proposed various computationally-efficient techniques for finding 'close to minimal' set covers, such as the **greedy heuristic** (Johnson, 1974); such algorithms can be adapted to the referring expression generation problem if desired.

## 2.3  Hearer Models

Dale (1992) suggests that the generation of referring expressions is governed by three principles, referred to as the principles of adequacy, efficiency and sensitivity; these are Gricean-like conversational maxims framed from the point of view of the specific task of generating referring expressions. The first two of these principles are primarily concerned with saying neither too much or too little: the principle of adequacy requires that a referring expression should contain enough information to allow the hearer to identify the referent, and the principle of efficiency requires that the referring expression should not contain unnecessary information. The principle of sensitivity, however, has a different concern: it specifies that the referring expression constructed should be sensitive to the needs and abilities of the hearer or reader. Accordingly, the definition of a distinguishing description specified in (3) above should really include a third component:

(4)    C3:   The hearer knows or can easily perceive that conditions C1 and C2 hold.

---

[3]See (Dale, 1992) for a referring expression generation algorithm that measures discriminatory power by assigning specific numerical values to attribute–value pairs based on their ability to distinguish the intended referent in context.

In other words, the hearer must realize that the distinguishing description matches the intended referent and none of the contrast objects, and ideally this realisation should not require a large perceptual or cognitive effort on the hearer's part. Thus, a distinguishing description should not mention a property that is not directly perceivable unless the hearer already knows the value of the attribute for the intended referent and for the contrast objects. It would be inappropriate, for example, to mention the manufacturer of a component in a distinguishing description if the hearer could only determine the manufacturer by dismantling the component. Appelt (1985b, page 6) gives a good example of a referring expression which satisfies, in our terms, conditions C1 and C2 but not C3: he describes a scenario where a speaker tells a hearer (whom she has just met on the bus) which bus stop to get off at by saying *Get off one stop before I do*. This may be a uniquely identifying description of the intended referent, but it is of little use without a supplementary offer to indicate the stop.

Furthermore, as Goodman (1986, page 285) points out, a distinguishing description should not use a very specific attribute value if the hearer is not familiar enough with this value to be able to distinguish it from other, similar, attribute values; there is no point, for example, in describing an object as *the magenta plug* if the hearer is not familiar enough with the meaning of *magenta* to be able to distinguish a magenta plug from a maroon plug.

A good referring expression generation algorithm should therefore be able to take into account what is known about the hearer's knowledge and perceptual abilities. This can be done at the simplest level by restricting the attributes mentioned in a referring expression to those which most human hearers are presumed to easily be able to perceive, such as colour and size. A more general solution is to allow the generation algorithm to issue appropriate queries to a hearer model at run-time; the algorithm we describe in Section 4 does this by calling a special UserKnows function.

## 2.4   Avoiding False Implicatures

Some distinguishing descriptions can be inappropriate referring expressions because they convey false implicatures to the human user. This is illustrated by the following examples of referring expressions that a speaker might use to request that a hearer sit by a table:

(5)    a.   Sit by *the table*.
       b.   Sit by *the brown wooden table*.

If the context was such that only one table was visible, and this table was brown and made of wood, utterances (5a) and (5b) would both be distinguishing descriptions that unambiguously identified the intended referent to the hearer; a hearer who heard either utterance would know where she was supposed to sit. However, a hearer who heard utterance (5b) in such a context might make the additional inference that it was important to the understanding of the discourse that the table was brown and made of wood; for, the hearer might reason, why else would the speaker include information about the table's colour and material that was not necessary for the reference task? If the sole purpose of the referring expression was to identify the table (and not to provide additional information about the table for some other purpose) then this inference is an example of a conversational implicature caused by a violation of Grice's maxim of Quantity (Grice, 1975).

Of course, in many contexts the speaker may indeed have additional communicative goals beyond simple identification. For example, if the speaker has the additional communicative goal of warning the hearer not to put her elbows on the table, it is perfectly appropriate to generate a referring expression such as:[4]

(6)     Sit by *the newly painted table*.

In this case, the hearer will (hopefully) make the inference that she shouldn't touch the table, but since the speaker *intended* for her to make this inference, this is not a *false* implicature. The point is that the referring expression should not lead a hearer to make *unintended* inferences, especially if these inferences are not true.

In order to simplify the analysis, in this paper we restrict our discussion to referring expressions that are only intended to identify an object, and have no other communicative purpose; we do *not* examine situations where a speaker has non-identificational communicative goals such as warning a hearer not to touch a table. A complete model of the generation of referring expressions would of course need to consider more general communicative goals.

### 2.4.1 The Gricean Maxims

We base our analysis of unwanted inferences on Grice's maxims of conversational implicature, which are listed in Figure 1. In the context of the referring expression generation task, Grice's maxims can be interpreted as follows:

**Quality:** A referring expression must be an accurate description of the intended referent.

**Quantity:** A referring expression should contain enough information to enable the hearer to identify the object referred to, but not more information.

**Relevance:** A referring expression should not mention attributes that have no discriminatory power, and hence do not help distinguish the intended referent from the members of the contrast set.

**Manner:** A referring expression should be short whenever possible.[5]

---

[4]This example, and our appreciation of its relevance to the current discussion, are due to Bonnie Webber.

[5]This corresponds to the Brevity submaxim of the Manner maxim; the other Manner submaxims will not be discussed here.

---

*Maxim of Quality:* Try to make your contribution one that is true. More specifically:

1. Do not say what you believe to be false.
2. Do not say that for which you lack adequate evidence.

*Maxim of Quantity:*

1. Make your contribution as informative as is required (for the current purposes of the exchange).
2. Do not make your contribution more informative than is required

*Maxim of Relevance:* Be relevant.

*Maxim of Manner:* Be perspicuous. More specifically:

1. Avoid obscurity of expression.
2. Avoid ambiguity.
3. Be brief (avoid unnecessary prolixy).
4. Be orderly.

Figure 1: The Gricean Maxims (excerpted from [Gri75, page 65])

---

An additional source of conversational implicatures is the failure to use words that correspond to **basic-level** classes (Rosch, 1978; Cruse, 1977). Consider the referring expressions used in (7a) and (7b):

(7)      a.     Look at *the dog*.
           b.     Look at *the pit bull*.

In a context where there is only one dog present, the hearer would normally expect utterance (7a) to be used, since *dog* is a basic-level class for most native speakers of English. The use of utterance (7b) might implicate to the hearer that the speaker thought it was relevant that the animal was a pit bull and not some other kind of dog (perhaps because the speaker wished to warn the hearer that the animal might be dangerous); if the speaker had no such intention, she should avoid using utterance (7b), despite the fact that it fulfills the referential communicative goal.

Basic-level classes are only one example of the **lexical preferences** that human hearers have and that generation systems should obey if possible (Reiter, 1991). A general lexical preference rule can be stated as a fifth maxim of conversational implicature:

**Lexical Preference:** Use basic-level and other lexically preferred classes whenever possible.

| Interpretation | Theoretical complexity | Typical run-time (Simple algorithm) |
|---|---|---|
| Full Brevity (Dale, 1989) | NP-Hard | $\approx n_a{}^{n_l}$ |
| Greedy Heuristic (Dale, 1989) | polynomial | $\approx n_a n_d n_l$ |
| Local Brevity (Reiter, 1990a) | polynomial | $\approx n_a n_d n_l$ |
| Incremental Algorithm | polynomial | $\approx n_d n_l$ |

$n_a$ = the number of properties known to be true of the intended referent
$n_d$ = the number of distractors in the current context
$n_l$ = the number of attributes mentioned in the final referring expression.

Figure 2: Some interpretations of the maxims and their costs

## 3 What Does it Mean to Obey Grice's Maxims?

### 3.1 Four Computational Interpretations of the Maxims

A natural language generation system must be based on precise computational rules, not the vague principles presented in Section 2.4, and there are many possible computational interpretations of the Gricean maxims. In general terms, any distinguishing description that is used as a referring expression will automatically satisfy the Quality maxim (that is, it will be a truthful description of the intended referent) and the first half of the Quantity maxim (that is, it will provide enough information to identify the intended referent). The remaining requirements can be summarized as follows:

- The referring expression should not include unnecessary information (the Maxim of Quantity).
- The referring expression should only specify properties that have some discriminatory power (the Maxim of Relevance).
- The referring expression should be short (the Maxim of Brevity).
- The referring expression should use basic-level and other lexically preferred classes whenever possible (Lexical Preference).

Note that the Brevity requirement in a sense subsumes the Quantity and Relevance requirements, since a referring expression that contains unnecessary or irrelevant information will also probably be longer than required. Given their rather vague definitions, it is not too surprising that the different maxims overlap and interact.

We discuss below four possible computational interpretations of these requirements, which we refer to as the Full Brevity Interpretation, the Greedy Heuristic Interpretation, the Local Brevity Interpretation, and the Incremental Algorithm Interpretation. The first three interpretations have been proposed in previous research; the last corresponds to the algorithm presented in the present paper.

The computational cost of generating referring expressions that satisfy each of these interpretations is summarized in Figure 2. These costs are stated in terms of the following parameters:

$n_a$ = the number of attributes that are available to be used in a referring expression (that is, the number of properties known to be true of the intended referent);

$n_d$ = the number of distractors in the current context;

$n_l$ = the number of attributes mentioned in the final referring expression.

For instance, recall the example discussed in Section 2.2; the task is to create a referring expression for Object1 in a context that also includes Object2 and Object3, where these objects possess the following properties:

- Object1: ⟨type, dog⟩, ⟨size, small⟩, ⟨colour, black⟩
- Object2: ⟨type, dog⟩, ⟨size, large⟩, ⟨colour, white⟩
- Object3: ⟨type, cat⟩, ⟨size, small⟩, ⟨colour, black⟩

In this situation, the values of the cost parameters are

$n_a$ = 3 (type, size, and colour);

$n_d$ = 2 (Object2, Object3); and

$n_l$ = 2 (⟨type,dog⟩, ⟨colour,black⟩).

Of course, this example is very simplistic. In more realistic situations, $n_l$, the number of attributes mentioned in the referring expression, will still typically be fairly small, but may reach four or five in some circumstances (for example, *the small black furry dog*, or *the upside-down scratched white porcelain cup*). However, in some situations the number of distractors, $n_d$, can be 50 or more (for example, consider identifying a particular book on a bookshelf, or a particular switch on a complex control panel); and $n_a$, the number of attributes that are available to be used in the referring expression, can easily reach 10 or 20, since almost any visually perceivable attribute can be used to identify a physical object (that is, not just colour and size, but also material, shape, orientation, texture, and so on).

### 3.1.1 The Full Brevity Interpretation

Dale (1989; 1992) suggested that a generation system should generate the *shortest possible* referring expression: that is, a referring expression that mentions as few attributes as possible. We will refer to this as the **Full Brevity Interpretation** of the maxims. A referring expression that satisfies the Full Brevity Interpretation will automatically satisfy a very strict interpretation of the Quantity, Relevance, and Brevity maxims. Unfortunately, finding the shortest possible referring expression is an NP-hard task (as discussed in Section 2.2), and this interpretation is therefore probably too computationally expensive to use in a natural language generation system operating in a domain of any real complexity.

For readers who are uncomfortable with the concept of NP-Hardness, it may be worthwhile to examine the computational expense of a specific algorithm that generates referring expressions that satisfy Full Brevity. The most straightforward such algorithm simply does an exhaustive search: it first checks if any one-component referring expression is successful, then checks if any two-component referring expression is successful, and so on. The expected run-time of this algorithm is roughly:[6]

$$\sum_{i=1}^{n_l} \frac{n_a!}{i!(n_a-i)!}$$

If $n_a \gg n_l$, this will be of order $\approx n_a^{n_l}$.

For the example problem discussed in Sections 2.2 and 3.1, $n_a$ is 3 and $n_l$ is 2, so the straightforward brevity algorithm will take only 6 steps to find the shortest description. In more realistic situations, however, things get much worse. For example:

- Suppose $n_l = 3$, and $n_a = 10$: then, 175 descriptions must be checked.
- Suppose $n_l = 4$, and $n_a = 20$: over 6000 descriptions must be checked.
- Suppose $n_l = 5$, and $n_a = 50$: over 2,000,000 descriptions must be checked.

A straightforward implementation of the Full Brevity Interpretation, then, seems prohibitively expensive in at least some circumstances. Since finding the shortest description is NP-Hard, it seems likely (existing complexity-theoretic techniques are too weak to prove such statements) that all algorithms for finding the shortest description will have similarly bad performance 'in the worst case'. It is possible that there exist algorithms that have acceptable performance in almost all 'realistic' cases; any such proposed algorithm, however, should be carefully analyzed to determine those circumstances in which it will fail to find the shortest description or in which it will take exponential time to run. Similarly, systems could be built that used the Full Brevity interpretation in simple contexts (low $n_a$ and $n_l$) but switched to a different interpretation in complex situations; this again should be carefully explained to potential users.

Any discussion of the impact of the above analysis for human generation of referring expressions must also be hedged with caveats. It is possible that the brain's parallelized architecture makes feasible algorithms that would be impractical on a sequential computer; there may also be psychological limitations on the number of properties and distractors that the human language processor is capable of considering, which would add an extra complication to the discussion. The available evidence, however (see Section 3.2.1), suggests that humans do not use the Full Brevity interpretation, at least in the experimental contexts that have been tested to date.

### 3.1.2 The Greedy Heuristic Interpretation

Dale (1989; 1992) proposed a referring expression generation algorithm that was essentially a variant of Johnson's greedy heuristic for minimal set cover (Johnson, 1974); we will refer to this as the **Greedy Heuristic Interpretation** of the maxims.[7] A version of this algorithm is shown in Figure 3. Dale also suggested meeting the lexical preference criteria by only using basic-level attribute values.

---

[6]This is just the number of potential descriptions that must be checked. The cost of checking each potential description to see if it is a distinguishing description is assumed to be small, since in most cases the system will probably quickly determine that the candidate description does not rule out all of the distractors.

[7]Appelt (1985a, pages 117–118) also suggested that referring expressions should be generated by a heuristic approximation to Full Brevity, although he did not specify any particular algorithm.

Let $L$ be the set of properties to be realised in our description; let $P$ be the set of properties known to be true of our intended referent $r$ (we assume that $P$ is non-empty); and let $C$ be the set of distractors (the contrast set). The initial conditions are thus as follows:

- $C = \{\langle \textit{all distractors} \rangle\}$;
- $P = \{\langle \textit{all properties true of } r \rangle\}$;
- $L = \{\}$

In order to describe the intended referent $r$ with respect to the contrast set $C$, we do the following:

1. Check Success:

   **if** $|C| = 0$ **then** return $L$ as a distinguishing description
   **elseif** $P = \emptyset$ **then** fail
   **else goto** Step 2.

2. Choose Property:

   **for each** $p_i \in P$ **do**: $C_i \leftarrow C \cap \{x|p_i(x)\}$
   Chosen property is $p_j$, where $C_j$ is the smallest set.
   **goto** Step 3.

3. Extend Description (wrt the chosen $p_j$):

   $L \leftarrow L \cup \{p_j\}$
   $C \leftarrow C_j$
   $P \leftarrow P - \{p_j\}$
   **goto** Step 1.

Figure 3: An algorithm for the Greedy Heuristic

This interpretation is much easier and quicker to implement than the Full Brevity interpretation discussed above, but such an algorithm will not in all cases produce the referring expression that uses the fewest possible attributes. Suppose, for example, we have seven cups, with internal names Object1 through Object7 and with the following properties:

- Object1: ⟨size, large⟩, ⟨colour, red⟩, ⟨material, plastic⟩
- Object2: ⟨size, small⟩, ⟨colour, red⟩, ⟨material, plastic⟩
- Object3: ⟨size, small⟩, ⟨colour, red⟩, ⟨material, paper⟩
- Object4: ⟨size, medium⟩, ⟨colour, red⟩, ⟨material, paper⟩
- Object5: ⟨size, large⟩, ⟨colour, green⟩, ⟨material, paper⟩
- Object6: ⟨size, large⟩, ⟨colour, blue⟩, ⟨material, paper⟩
- Object7: ⟨size, large⟩, ⟨colour, blue⟩, ⟨material, plastic⟩

The algorithm as specified first selects *plastic*; then selects *large* or *red*; and finally selects *red* or *large* (whichever of the two was not selected in the second step). The result, once the head noun has been added, is the noun phrase *the large red plastic cup*; however, the true minimal description is *the large red cup*.

From a computational perspective, the greedy heuristic requires $n_l$ passes through the problem, and each pass requires checking each of the $n_a$ potential attributes to determine how many of the $n_d$ distractors they rule out. The total running time is therefore of the order $\approx n_a n_d n_l$.

### 3.1.3 The Local Brevity Interpretation

Reiter (1990a) proposed that referring expression generation systems should generate utterances that were maximal under the following preference rules:

**No Unnecessary Components:** all components of a referring expression must be necessary to fulfill the referential goal. For example, *the small black dog* is not acceptable if *the black dog* is a distinguishing description, since this means *small* is an unnecessary component.

**Local Brevity:** it should not be possible to produce a shorter referring expression by replacing a set of existing components by a single new component. For example, *the sleeping female dog* should not be used if *the small dog* is a distinguishing description, since the two modifiers *sleeping* and *female* can be replaced by the single modifier *small*.

**Lexical Preference:** basic-level and other lexically-preferred words (Reiter, 1991) should be used whenever possible.

We will refer to Reiter's interpretations of the maxims as the **Local Brevity Interpretation**. The term 'component' in Reiter's preference rules can be interpreted either semantically (that is, each attribute–value pair is a component) or syntactically (for example, each open-class word in the surface form is a component). Reiter points out that defining components in syntactic terms leads to computationally very expensive implementations, and hence suggests using a semantic definition.

If a semantic interpretation of 'component' is used and the lexical preference relation meets certain constraints (Reiter, 1990a) then a referring expression that meets Reiter's conditions can be constructed by a simple iterative algorithm. This algorithm starts with an initial distinguishing description, and checks to see if a new distinguishing description can be formed from the current description by applying any of the preference rules (that is, by removing an attribute, replacing two or more attributes by a single attribute, or replacing a value by a lexically-preferred value). If so, the algorithm iterates with the new description; if not, the current description must satisfy the preference rules, and hence can be uttered.

In computational complexity terms, the algorithm always runs in polynomial time. The algorithm requires an initial distinguishing description; if the Greedy Heuristic is used to generate the initial description, this will require time $\approx n_a n_d n_l$. The iterative improvement steps can also be executed in this order of time,[8] so the total cost of the algorithm, including both

---

[8] Checking the Local Brevity rule is the most time-consuming operation; one simple way of doing this is to replace in turn each of $n_l$ attributes in the current description by each of the $(n_a - n_l)$ attributes not mentioned in the description, and then check if this forms a distinguishing description; if so, the system then needs to

forming the initial description with the Greedy Heuristic and performing the optimization passes, will also be of order $\approx n_a n_d n_l$.

### 3.1.4 Incremental Algorithm Interpretation

The **Incremental Algorithm Interpretation** is embodied in the algorithm presented in Section 4 of this paper. This algorithm simply sequentially iterates through a (task-dependent) list of attributes, adding an attribute to the description being constructed if it rules out any distractors that have not already been ruled out, and terminating when a distinguishing description has been constructed. This algorithm has an expected typical-case run-time of about $n_d n_l$; that is, it is faster than the above algorithms by roughly a factor of $n_a$, the number of attributes that are available to be included in a referring expression. This is a direct consequence of the fact that the Incremental Algorithm does *not* attempt to look for 'optimal' attributes that rule out as many distractors as possible, but rather simply iterates through the list of available attributes in a fixed order.

## 3.2 What Do People Do?

Human speakers presumably face at least some of the computational limitations that computer natural language generation systems face, so it is useful to examine the kinds of referring expressions they generate. Human hearers also expect to encounter the kind of referring expressions generated by human speakers, so a natural language generation system may be easier to use if it generates referring expressions which are similar to those generated by human speakers.

### 3.2.1 Psychological Evidence

We have looked in some detail at the psychological literature on the human generation of referring expressions (for example, Ford & Olson, 1975; Whitehurst, 1976; Sonnenschein, 1985; Pechmann, 1989; Levelt (1989, pages 129–134) provides a useful summary of much of this work). Two salient points that emerge from this research are:

**Observation 1:** Human speakers in many cases include unnecessary modifiers in the referring expressions they construct.

**Observation 2:** Human speakers can begin to utter a referring expression before they have finished scanning the set of distractors.

Observation 1 is supported by evidence in all of the studies cited above. For example, in a typical experiment, a subject is shown a picture containing a white bird, a black cup, and a white cup, and is asked to identify the white bird; in such cases subjects generally produce

---

determine if any further attributes can also be removed, thus producing a shorter distinguishing description. If we assume that in most cases the modified descriptions will not be distinguishing descriptions, so that the second stage is rarely necessary, each optimization pass will require on the order of $n_a n_d n_l$ steps. The Greedy Heuristic is fairly efficient, so in the great majority of cases only one or two optimization iterations will be necessary; at least one such must always be done, to verify that no preferred description exists.

the referring expression *the white bird*, even though the simpler form *the bird* would have been sufficient.[9]

Observation 2 is supported by eye-tracker experiments carried out by Pechmann (1989); Pechmann's experiments are similar to the one described above, except that the movements of the subject's eyes are recorded.

The two points may be connected, since an algorithm that begins outputting a referring expression before it has completed scanning the set of distractors can easily produce referring expressions that contain unnecessary modifiers. In the above example, for instance, one could imagine a speaker scanning the black cup and deciding to use the adjective *white* to rule it out, and then scanning the white cup and deciding to use the noun *bird* in order to rule it out; *bird* also rules out the first distractor (the black cup) and hence its presence makes *white* unnecessary, but the speaker cannot 'unsay' *white* if she has already uttered this word. In McDonald's terminology (McDonald, 1981; 1983), Pechmann's data supports the hypothesis that human speakers generate referring expressions incrementally and indelibly; they make certain decisions about content before they have finished computing the entire referring expression, and do not backtrack and undo these decisions if they later turn out to produce a non-optimal utterance.

There are other possible explanations for the inclusion of unnecessary modifiers. For example, the speaker may wish to make it as easy as possible for the hearer to identify the object, and believe that including extra modifiers may be helpful in this regard. Thus, for instance, the speaker may believe that it is easier for the hearer to identify an object if it is referred to as a *white bird* rather than simply as a *bird*, since colour may be more immediately perceptible than genus.

Another possible explanation is that speakers may in some cases use precompiled 'reference scripts' instead of computing a referring expression from scratch; such reference scripts would specify a set of attributes that are included as a group in a referring expression, even if some members of the group turn out to have no discriminatory power in the current context. For example, the speaker might have noticed that in past situations of a similar nature an effective referring expression could be formed with the `colour` and `type` attributes, and hence simply include these attributes as a group in the referring expression, without checking to see whether each individual member of the group was in fact necessary in the current context.

The first two explanations would account for the inclusion of modifiers that have a non-null *RulesOut* value, that is, that rule out at least one distractor (such a modifier is unnecessary if the distractors it rules out are also ruled out by other components of the referring expression). The reference-script explanation, in contrast, might justify the inclusion of modifiers that have no discriminatory power, that is, modifiers that do not rule out any distractor (for example, *white* would be such a modifier if all the objects in the contrast set shared with the intended referent the property ⟨colour, white⟩). Ford and Olson (1975) have observed that children do not seem to include modifiers that have no discriminatory power, but this was a *post hoc* comment, not a properly tested hypothesis, and in any case may not generalize to adult

---

[9]Merrill Garrett (personal communication) observes that experiments of this kind may not be valid for our purposes, since the experimental subject may assume that the stimulus carries only properties which are relevant for the purposes of identification, and the properties being explored by the experimenter may therefore take on an unnatural degree of salience. As always in such contexts, we must be wary of assuming that the experimental results tell us anything about behaviour in more natural situations.

speakers. Further psycholinguistic research is needed to clarify this issue.

Of the four computational interpretations of the Gricean maxims presented in Section 3.1, only the Greedy Heuristic and Incremental Algorithm interpretations would produce behaviour (1) above: Full Brevity and Local Brevity would never include unnecessary modifiers (for example, they would never produce *white bird* when *bird* would suffice). The Greedy Heuristic (as well as the Full Brevity and Local Brevity interpretations) is incompatible with Observation 2: it requires the set of distractors to be considered as a whole, and cannot produce partial output when the set of distractors has only been partially scanned. Thus, of the four interpretations proposed in Section 3.1, the Incremental Algorithm seems closest to what human speakers do (and indeed we designed this algorithm in the light of Observations 1 and 2).

The fact that human speakers include redundant information in their referring expressions suggests that, at least in some situations, and in contrast to our discussion of example (5) in Section 2.4, human hearers will not make unwanted implicatures from the presence of the unnecessary modifiers. We currently have insufficient linguistic and psycholinguistic knowledge to let us predict exactly when human hearers will or will not make conversational implicatures from 'violations' of the maxims, but clearly it is unnecessary for a natural language generation system to ensure that its output *never* contains redundant information.

### 3.2.2 Some Additional Hypotheses

Unfortunately, the psychological data available does not provide enough information to tightly constrain the computational models we might build. For working systems, there are further questions to which we require answers. Ultimately the relevant data may be obtained from experimentation and from corpus analysis, but in the interim we will make do with a number of plausible hypotheses. These hypotheses are partially based on a preliminary analysis of a transcript of a dialogue between two humans assembling a piece of garden furniture.[10]

For our purposes, we are particularly interested in questions of modifier choice; if a distinguishing description can be formed by adding any one of several modifiers to a head noun, which modifier should be used? In particular, we asked the following questions:

1. Faced with a choice, which attribute should be used? Suppose *the screw* is not a distinguishing description, but each of the expressions *the small screw*, *the black screw*, and *the screw made in Korea* is; which attribute will be chosen?
2. Is it preferable to add a modifier or to use a more specific head noun? For example, is it better to say *the small screw* or *the woodscrew*?[11]
3. Should relative or absolute adjectives be used? For example, is it better to say *the long screw* or *the three-inch screw*?

The following hypotheses seem intuitively plausible for situations involving spoken, face-to-face language:

---

[10]The transcript was made by Phil Agre and John Batali from a videotape taken by Candy Sidner; we acknowledge their contribution in allowing us to use it.

[11]There is also the question of whether one should use a modifier–noun combination or a more specific noun with the same semantics: consider *unmarried man* as compared to *bachelor*. We will not consider this question here.

1. Human speakers prefer to use adjectives that describe easily perceptible properties such as size, shape, or colour in referring expressions. In the transcript, for example, referring expressions such as *long screw*, *little dowel*, and *big bolt* were common; no referring expression mentioned a property that was not visually perceptible.

2. Human hearers sometimes have trouble determining if an object belongs to a specialized class. In the transcript we examined, for example, when the written instructions used specialized terms like *the lag screw*, in many cases one of the human assemblers had to explain to the other assembler that this meant *the long screw*. If there is any doubt about a human hearer's ability to distinguish objects that belong to a specialized class, it is better for a generation system to add an explicit modifier to the referring expression, and thus, for example, produce *the long screw* instead of *the lag screw*.

3. Human speakers seem to prefer to use relative adjectives, and human hearers seem to have less trouble understanding them. However, human-written instructional texts sometimes use absolute adjectives instead of relative ones; this may be a consequence of the fact that writers cannot predict the context in which their text will be read, and hence how readers will interpret relative adjectives. This was very evident in the transcript. The written assembly instructions in all cases used absolute modifiers, for example, $3\frac{1}{4}$" *bolt*.[12] In contrast, the human assemblers, unless they were reading or discussing the written instructions, in all cases used relative modifiers, such as *the long bolt*.

Of course, an extensive corpus analysis across a broad range of discourse types would be required in order to validate these observations, but for the moment they constitute a plausible way of constraining further our choice of computational models.

## 3.3 Evaluation of the Different Interpretations of the Maxims

In the light of the insights gained from our analyses of the relevant psychological research, and the hypotheses presented above, we are now in a position to evaluate the four different strategies for creating computational interpretations of the Gricean maxims presented in Section 3.1. We are particularly interested here in the way in which the Brevity submaxim is interpreted.

**The Full Brevity Interpretation:** This interprets Brevity in a straightforward and literal way, and ignores computational difficulties. So, 'be brief' becomes 'use the shortest possible referring expression', which is an NP-Hard task and thus very expensive to implement.

**The Greedy Heuristic Interpretation:** This acknowledges the computational difficulty of implementing a literal version of Brevity, and uses a well-known approximation algorithm to find a distinguishing description that hopefully is not much longer than the shortest possible referring expression.

**The Local Brevity Interpretation:** This acknowledges the computational difficulty of implementing a literal version of Brevity, and uses a variant of Brevity that can be implemented with a polynomial time algorithm. One major difference between the Local

---
[12]The symbol " is an abbreviation for *inch*.

Brevity and Greedy Heuristic interpretations is that it is possible to state in declarative terms what properties a referring expression must have in order to satisfy the Local Brevity Interpretation ('it should not be possible to produce a shorter referring expression by replacing a set of existing components by a single new component'), while the Greedy Heuristic Interpretation is characterised purely in algorithmic terms—its output cannot be characterized in declarative terms except as something 'close to' the shortest possible referring expression. In other words, the Local Brevity Interpretation is a *declarative* computationally tractable approximation to Full Brevity, while the Greedy Heuristic Interpretation is an *algorithmic* computationally tractable approximation to Full Brevity.

**The Incremental Algorithm Interpretation:** This does not impose any explicit rigorous Brevity constraint; instead, it attempts to mimic what human speakers do (as far as this is known) and generates appropriate utterances by virtue of this mimicry. This results in a simple and fast algorithm that does not have any explicit brevity requirement coded into it.

In short, the above models represent four approaches to creating computational interpretations of the Gricean maxims:

- Use a straightforward and literal interpretation of the maxims (the Full Brevity Interpretation).

- Use an algorithmic approximation to a straightforward interpretation (the Greedy Heuristic Interpretation).

- Use a declarative approximation to a straightforward interpretation (the Local Brevity Interpretation).

- Determine how human speakers generate text, and build an algorithm based on these observations (the Incremental Algorithm Interpretation); the maxims are not explicitly taken into consideration, but presumably an algorithm that mimics human speakers will obey the maxims as much as human speakers do.

The straightforward literal interpretation of the maxims is too computationally expensive to use in a practical system. In previous papers we have argued for the approximation approaches, claiming that a computer system should get as close as computationally feasible to the literal interpretation of the maxims. We now believe, however, that even these algorithms may be unnecessarily complicated and expensive in computational terms; if human speakers can 'get away' with producing referring expressions that do not conform literally to the maxims, there is no reason why computer natural language generation systems should be forced to conform to a literal reading. Indeed, the behaviour of human speakers here may be a good reason for computer systems *not* to adopt a literal reading.[13]

The argument can be made that psychological realism is not the most important consideration for developing algorithms for embodiment in computational systems; in the current context, the goal of such algorithms should be to produce referring expressions that human hearers will understand, rather than referring expressions that human speakers would utter. The fact (for example) that human speakers include redundant modifiers in referring expressions does not mean that natural language generation systems are also required to include such modifiers; there is nothing in principle wrong with building generation systems that perform more optimizations of their output than human speakers. On the other hand, if such beyond-human-speaker optimizations are computationally expensive and require complex algorithms, they may not be worth performing; they are clearly unnecessary in some sense, after all, since human speakers do not perform them.

---

[13] Again, we must add a caveat with regard to the psychological evidence: it is quite possible that people use a variety of different strategies as the situation demands. Ultimately, at most we have evidence for human behaviour in situations which are sufficiently like the experimental contexts discussed in Section 3.2.1.

One could even argue that an algorithm based on psycholinguistic observations of human speakers may in fact be superior to one that attempts to interpret the maxims as strictly as (computationally) possible. This would be justified if one believed that the Gricean maxims were simply an approximation to the general principle of 'if a speaker utters an unexpected utterance, the hearer may try to infer a reason for the speaker's failure to use the expected utterance'; under this perspective, a system that imitated human behaviour would be more likely to generate 'expected' utterances than a system that simply tried to obey general principles such as brevity, relevance, and so on. There is as yet insufficient data to determine whether the Gricean maxims are in fact simply approximations to this general principle; psycholinguistic experiments to test this hypothesis would be very useful.

One general lesson that can be drawn from the above analysis of the task of generating referring expressions is that the Gricean maxims should not be interpreted too literally (as is done, for example, in the Full Brevity Interpretation); no doubt Grice himself would have been the first to say this.

# 4 The Algorithm

Based on the above considerations, we have developed a new algorithm for generating referring expressions. This algorithm is simpler and faster than the algorithms proposed in (Dale, 1989; Reiter, 1990a) because it performs much less length-oriented optimization of its output; we now believe that the level of optimization suggested in (Dale, 1989; Reiter, 1990a) was unnecessary and psycholinguistically implausible. The algorithm described here has been implemented as part of a larger natural language generation system called IDAS (Reiter et al, 1992).

## 4.1 Assumptions about the Knowledge Base

One of our main concerns is the development of algorithms which are genuinely portable from one application to another and from one domain to another. With this goal in mind, our algorithm is intended to be reasonably domain-independent. We do, however, make some assumptions about the structure of the host system's underlying knowledge base, and require that certain interface functions be provided.

In particular, we assume that:

- Every entity is characterised in terms of a collection of **attributes** and their **values**.

As noted earlier, an attribute–value pair is what is sometimes thought of as a property; an example is ⟨colour, red⟩.

- Every entity has as one of its attributes some **type**. This is a special attribute that corresponds to the kinds of properties that are typically realized by head nouns; an example is ⟨type, dog⟩.
- The knowledge base may organize some attribute values in a subsumption taxonomy (for example, as is done in KL-ONE (Brachman & Schmolze, 1985) and related KR systems). Such a taxonomy might record, for example, that animal subsumes dog, and that red subsumes scarlet. Although in many systems only what we call the type attribute would have its possible values arranged as a taxonomy, for generality our algorithm permits a taxonomy of values for any attribute.

We require that the following interface functions be provided:

MoreSpecificValue(object, attribute, value) returns a new value for attribute where that value is more specific than value (that is, it is a child of value in the taxonomy of values for attribute), and subsumes the most specific value of attribute known for object from the system's point of view. If no more specific value is available, the function returns nil.

BasicLevelValue(object, attribute) returns the basic-level value of an attribute of an object, from the point of view of the current user. For example, BasicLevelValue(Garfield, type) might be cat. The knowledge representation system should in principle allow different basic-level classes to be specified for different users (Rosch, 1978; Reiter, 1991).

UserKnows(object, attribute-value-pair) returns true if the user knows or can easily determine (for example, by direct visual perception) that the attribute–value pair applies to the object; false if the user knows or can easily determine that the attribute–value pair does *not* apply to the object; and unknown otherwise. For example, if object $x$ had the property ⟨type, chihuahua⟩, and the user was capable of distinguishing dogs from cats, then UserKnows($x$, ⟨type, dog⟩) would be true, while UserKnows($x$, ⟨type, cat⟩) would be false. If the user was not, however, capable of distinguishing different breeds of dogs, and had no prior knowledge of $x$'s breed, then UserKnows($x$, ⟨type, chihuahua⟩) and UserKnows($x$, ⟨type, poodle⟩) would both return unknown, since the user would not know or be able to easily determine whether $x$ was a chihuahua, poodle, or some other breed of dog.

Finally, we assume that the global variable PreferredAttributes contains the attributes that human speakers and hearers prefer (for example, type, size, shape, and colour in the assembly task transcript mentioned in Section 3.2.2; and type, colour, and label in the machinery domain of Section 4.5). These attributes should be listed in order of preference, with the most preferred attribute first. The elements of this list and their order will vary with the domain, and will typically be determined by empirical investigation. It is this element that provides the real interface to domain portability.

## 4.2 Inputs and Outputs

In order to construct a reference to a particular entity, the host system must provide:

```
(X / Dog
   :determiner definite
   :relations ((Y / Colour
                  :domain X
                  :range (Z / Black))))
```

Figure 4: The SPL term corresponding to {⟨type,dog⟩,⟨colour,black⟩}

$$\begin{bmatrix} \text{index: } x \\ \text{status: } \begin{bmatrix} \text{given: } + \\ \text{unique: } + \end{bmatrix} \\ \text{spec: } \begin{bmatrix} \text{agr: } \begin{bmatrix} \text{countable: } + \\ \text{number: sg} \end{bmatrix} \\ \text{type: } \begin{bmatrix} \text{category: dog} \\ \text{properties: } \begin{bmatrix} \text{colour: black} \end{bmatrix} \end{bmatrix} \end{bmatrix} \end{bmatrix}$$

Figure 5: The recoverable semantic structure corresponding to {⟨type,dog⟩,⟨colour,black⟩}

- a symbol corresponding to the intended referent; and
- a list of symbols corresponding to the members of the contrast set (that is, the other entities in focus, besides the intended referent).

The algorithm returns a list of attribute–value pairs that correspond to the semantic content of the referring expression to be realized. This list can then be converted into an SPL (Kasper, 1989) term, as is done in the IDAS implementation; it can also be converted into a **recoverable semantic structure** of the kind used in Dale's EPICURE system (Dale, 1989; 1992). For example, the distinguishing description corresponding to the attribute-value list {⟨type,dog⟩,⟨colour,black⟩} would be converted into the SPL term shown in Figure 4 and the recoverable semantic structure shown in Figure 5.

## 4.3 The Algorithm

In general terms, the algorithm iterates through the attributes in PreferredAttributes. For each attribute, it checks if specifying a value for that attribute would rule out at least one member of the contrast set that has not already been ruled out; if so, this attribute is added to the set of attributes to be used in the referring expression, with a value that is known to the user, rules out as many contrast set members as possible, and, subject to these constraints, is as close as possible to the basic-level value. The process of adding attribute–value pairs continues until a referring expression has been formed that rules out every member of the contrast set. There is no backtracking; once an attribute–value pair has been added to the referring expression, it is not removed even if the addition of subsequent attribute–value pairs make it unnecessary. A head noun (that is, a value for the type attribute) is always included,

---

$\boxed{\mathsf{MakeReferringExpression}(r,\ C,\ P)}$

$L \leftarrow \{\}$
**for** each member $A_i$ of list $P$ **do**
   $V = \mathsf{FindBestValue}(r,\ A_i,\ \mathsf{BasicLevelValue}(r,\ A_i))$
   **if** $\mathsf{RulesOut}(\langle A_i,\ V \rangle) \neq \mathsf{nil}$
   **then** $L \leftarrow L \cup \{\langle A_i, V \rangle\}$
       $C \leftarrow C - \mathsf{RulesOut}(\langle A_i,\ V \rangle)$
   **endif**
   **if** $C = \{\}$ **then**
     **if** $\langle \mathsf{type},\ X \rangle \in L$ for some $X$
       **then return** $L$
       **else return** $L \cup \{\langle \mathsf{type},\ \mathsf{BasicLevelValue}(r,\ \mathsf{type}) \rangle\}$
       **endif**
   **endif**
**return** failure

$\boxed{\mathsf{FindBestValue}(r,\ A,\ \textit{initial-value})}$

**if** $\mathsf{UserKnows}(r,\ \langle A,\ \textit{initial-value} \rangle) = \mathsf{true}$
**then** $\textit{value} \leftarrow \textit{initial-value}$
**else** $\textit{value} \leftarrow \mathsf{no\text{-}value}$
**endif**
**if** $(\textit{more-specific-value} \leftarrow \mathsf{MoreSpecificValue}(r,\ A,\ \textit{value})) \neq \mathsf{nil}\ \wedge$
   $(\textit{new-value} \leftarrow \mathsf{FindBestValue}(A,\ \textit{more-specific-value})) \neq \mathsf{nil}\ \wedge$
   $(|\mathsf{RulesOut}(\langle A,\ \textit{new-value} \rangle)| > |\mathsf{RulesOut}(\langle A,\ \textit{value} \rangle)|)$
**then** $\textit{value} \leftarrow \textit{new-value}$
**endif**
**return** $\textit{value}$

$\boxed{\mathsf{RulesOut}(\langle A,\ V \rangle)}$

**if** $V = \mathsf{no\text{-}value}$
**then return** nil
**else return** $\{x : x \in C \wedge \mathsf{UserKnows}(x,\ \langle A,\ V \rangle) = \mathsf{false}\}$
**endif**

Figure 6: The Algorithm

---

even if it has no discriminatory power (in which case the basic level value is used); other attribute values are only included if, at the time they were under consideration, they ruled out some distractors that had not been ruled out by previously considered attributes.

More precisely, the algorithm is as shown in Figure 6. Here, $r$ is the intended referent, $C$ is the contrast set, $P$ is the list of preferred attributes, and $L$ is the list of attribute–value pairs returned.[14]

MakeReferringExpression is the top level function. This returns a list of attribute–value pairs that specify a referring expression for the intended referent. The attributes are tried in the order specified in the PreferredAttributes list, and a value for type is always included, even if type has no discriminatory power.

---

[14]For simplicity of exposition, the algorithm as described here returns failure if it is not possible to rule out all the members of the contrast set. A more robust algorithm might attempt to pursue other strategies here, for example, generating a referring expression of the form *one of the Xs* (see Dale, 1992), or modifying the contrast set by adding navigation information (see Section 5.1.1).

FindBestValue takes an attribute and an initial value; it returns a value for that attribute that is subsumed by the initial value, accurately describes the intended referent (that is, subsumes the value the intended referent possesses for the attribute), rules out as many distractors as possible, and, subject to these constraints, is as close as possible in the taxonomy to the initial value. It does this by considering successively more specific values for the attribute, stopping when the most specific value known to the user is reached, and choosing as the best value that value which rules out most elements of the contrast set. When two values of different specificity rule out the same number of elements, the least specific value is chosen. If a value that the user knows to be true cannot be found, the function returns no-value.

RulesOut takes an attribute–value pair and returns the elements of the set of remaining distractors that are ruled out by this attribute–value pair.

## 4.4 An Example

To see how this algorithm works, we will consider an example in detail. Suppose the task is to create a referring expression for Object1 in a context that also includes Object2 and Object3, where these objects are known by the system to have the following properties:[15]

- Object1: ⟨type, chihuahua⟩, ⟨size, small⟩, ⟨colour, black⟩
- Object2: ⟨type, chihuahua⟩, ⟨size, large⟩, ⟨colour, white⟩
- Object3: ⟨type, siamese-cat⟩, ⟨size, small⟩, ⟨colour, black⟩

In other words, $r$ = Object1 and $C$ = {Object2, Object3}. Assume that $P$ = {type, colour, size, ...}.

When MakeReferringExpression is called in this context, it initializes $L$ to the empty set. FindBestValue is then called with $A$ = type, and *initial-value* set to the basic-level type of Object1, which, let us assume, is dog.

Assume UserKnows(Object1, ⟨type, dog⟩) is true, that is, the user knows or can easily perceive that Object1 is a dog. FindBestValue then sets *value* to dog, and examines the more specific values for type for the intended referent to see if any of them rule out more distractors than dog does. In this case, the only appropriate descendent of dog is chihuahua, but ⟨type, chihuahua⟩ does not have more discriminatory power than ⟨type, dog⟩ (both rule out {Object3}), so FindBestValue returns dog as the best value for the type attribute. MakeReferringExpression then verifies that ⟨type, dog⟩ rules out at least one distractor, and therefore adds this attribute–value pair to $L$, while removing RulesOut(⟨type, dog⟩) = {Object3} from $C$.

This means that the only remaining distractor in $C$ is Object2. MakeReferringExpression (after checking that $C$ is not empty) calls FindBestValue again with $A$ = colour (the second member of $P$). FindBestValue returns Object1's basic-level colour value, which is black, since no more specific colour term has more discriminatory power. MakeReferringExpression then adds ⟨colour, black⟩ to $L$ and removes RulesOut(⟨colour, black⟩) = {Object2} from $C$. $C$ is then empty, so the generation task is completed, and MakeReferringExpression returns {⟨type, dog⟩, ⟨colour, black⟩}, that is, a specification for the referring expression *the black dog*. Note that if $P$ had been {type, size, colour, ...} instead of {type, colour, size, ...}, MakeReferringExpression would have returned {⟨type, dog⟩, ⟨size, small⟩} instead; that is, the semantic content of the referring expression *the small dog*.[16]

---

[15] This example is the same as the example of Section 2.2, except that the non-basic-level types chihuahua and siamese-cat have been used instead of the basic-level terms dog and cat.

## 4.5 Implementation

The algorithm is currently being used within the IDAS system (Reiter et al 1992). IDAS is a natural language generation system that generates on-line documentation and help texts from a domain knowledge base and models of the user's task, the user's expertise level, and the discourse context. IDAS uses a KL-ONE-like knowledge representation system, with roles corresponding to attributes and fillers to values. The type attribute is implicit in the position of an object in the taxonomy, and is not explicitly represented. The MoreSpecificValue function is defined in terms of standard knowledge base access functions.

A knowledge-base author can specify explicit basic-level attribute values in IDAS user models, but IDAS is also capable of using heuristics to guess which value is basic-level. The heuristics are fairly simple (for example, 'use the most general value that is not in the upper-model (Bateman et al, 1990) and has a one-word realization'), but they seem to be fairly effective, at least in IDAS's current domain (complex electronic machinery). A PreferredAttributes list has been created for this domain by visual inspection of the equipment being documented; its first members are type, colour, and label.[17] The UserKnows function simply returns true if the attribute–value pair is accurate and false otherwise; this essentially assumes that the user can visually perceive the value of any attribute in PreferredAttributes, which may not be true in general.

The algorithm performs quite effectively in IDAS. In particular, the algorithm has proven to be useful because:

1. It is fast. The algorithm's run-time is linear in the number of distractors and independent of the number of potential attributes (see Section 3.1).
2. It allows human preferences and capabilities to be taken into consideration. The PreferredAttributes list, the preference for basic-level values, and the UserKnows function are all ways of biasing the algorithm towards generating referring expressions that use attributes and values that human hearers, with all their perceptual limitations, find easy to process, thus satisfying the principle of sensitivity (see Section 2.3).

Almost all referring expressions generated by IDAS contain a head noun and zero, one, or perhaps at most two modifiers; longer referring expressions are rare. The most important task of the algorithm is therefore to quickly generate easy-to-understand referring expressions in such simple cases; optimal handling of more complex referring expressions is less important, although the algorithm should be robust enough to generate something plausible if a longer referring expression is needed.

---

[16]Note that this same semantic content could be realized as *the dog which is small*. Again, we do not address here the issue of what constraints are brought to bear in the mapping from semantic content to linguistic form, and assume for the moment that both realisations are equivalent.

[17]Label is the textual label that some switches, indicators, etc., have; for example, ⟨label,Power⟩ means that the entity in question has a *Power* label attached to it.

# 5 Conclusions and Future Work

We have presented four possible interpretations of the Gricean maxim of Brevity for the referring expression generation task, and argued that the simplest and fastest one may be the best to use, because it seems to be closest to what human speakers do. Based on our chosen interpretation, we have created an algorithm for the generation of referring expressions that is substantially simpler and faster than the algorithms we have proposed in previous work (Dale, 1989; Reiter, 1990a). The algorithm has been defined in terms of well-specified interface functions, and hence should be portable to most domains.

There are two general areas we are particularly interested in pursuing in future work; generating more complex referring expressions, and investigating different computational interpretations of the Gricean maxims in other natural language generation tasks.

## 5.1 More Complex Referring Expressions

### 5.1.1 Navigation

At the beginning of this paper, we stated that the purpose of the algorithm presented here is to distinguish an intended referent from all other entities that are in the context set, where this is taken to be the set of entities which are in the current focus of attention. An important question we need to address is what action should be taken if the intended referent is not in the current focus of attention.

Unfortunately, there is very little psycholinguistic data available on which to base a model of the generation of such referring expressions. However, we think it is useful to take the view that, in the general case, a referring expression contains two kinds of information: **navigation** and **discrimination**. Each descriptor used in a referring expression plays one of these two roles.

- Navigational, or **attention-directing** information, is intended to bring the intended referent into the hearer's focus of attention.
- Discrimination information is intended to distinguish the intended referent from other objects in the hearer's focus of attention; such information has been the subject of this paper.

Navigational information is not needed if the intended referent is already in the focus of attention. If it is needed, it typically takes the form, at least for physical objects, of locational information.[18] The IDAS system, for example, can generate referring expressions such as *the*

---

[18]Note that we are not implying here that objects which are outside of the focus of attention are necessarily

*black power supply in the equipment rack*. In this case, *in the equipment rack* is navigation information that is intended to bring the equipment rack and its components into the hearer's focus of attention, while *black power supply* is discrimination information that is intended to distinguish the intended referent from other members of the context set (for example, the white power supply that is also present in the equipment rack). Appelt's KAMP system (Appelt, 1985a; 1985b) was also capable of generating referring expressions that, in our terminology, included navigational information. For example, KAMP could generate the utterance *Remove the pump with the wrench in the tool-box*, where *the wrench in the tool-box* is a referring expression that uses *in the tool-box* as navigational information to tell the hearer where the tool can be found, and *wrench* as discrimination information that informs the hearer which tool from the tool-box should be used.

The navigation model currently implemented in IDAS is simplistic and not theoretically well-justified, and we believe that it would be difficult to implement the KAMP model in a computationally efficient manner. We hope to do further research on building a theoretically sound and computationally efficient model of navigation.

### 5.1.2 Relative Attribute Values

We suggested earlier that, at least in contexts where the intended referent is physically co-present for both speaker and hearer, the speaker will often prefer to use relative instead of absolute attribute values; for example, *small* instead of *one inch*. We make no claims here as to how people encode the properties of objects in such situations, although clearly in some situations the encoding must be in relative terms: we can tell the relative heights of two neighbouring mountains without having any idea as to their absolute heights. Similarly, in computational systems, knowledge bases sometimes explicitly encode relative attribute values (for example, ⟨size, small⟩), but this can cause difficulties when referring expressions need to be generated in different contexts; a one-inch screw, for example, might be considered to be small in a context where the other screws were all two-inch screws, but large in a context where the other screws were all half-inch screws.

From the point of view of constructing broad coverage, portable computational systems, a better solution is for the knowledge base to record absolute attribute values where they are available, and then for the generation algorithm to automatically convert absolute values to relative values, depending on the values that other members of the context set possess for this attribute.[19] Thus, the knowledge base might record that a particular screw had ⟨size, one-inch⟩, and the generation system would choose to call this screw *small* or *large* depending on the size of the other screws in the context set. We hope to do further research on determining how exactly this process should work.

---

outside of the hearer's field of vision: an object may be in the hearer's field of vision and still not in her focus of attention, and can be outside of her field of vision but still in her focus of attention

[19]Of course, if the generation system is a component in a larger intelligent system that also incorporates a vision component, it is quite possible that the system may find itself in situations where it is only able to construct representations that contain relative values for attributes.

## 5.2 Computational Interpretations of the Maxims in Other NLG Tasks

The principle that has emerged from our study of the referring expression generation task is that a simple and non-literal interpretation of the Gricean maxims is to be preferred: it is both faster in computational terms, and seems to be closer to what human speakers do when they construct referring expressions. We are very interested in performing similar studies of other generation tasks, to determine if the same principle applies in these tasks as well. Perhaps it may some day be possible to make a very general statement such as 'human speakers in general use very simple (in computational terms) interpretations of the maxims of conversational implicature, and hence computer natural language generation systems should also use such interpretations'; we do not as yet have sufficient evidence to support such a sweeping claim, but we intuitively believe that there is some truth to it, and hope to be able to provide more evidence to support this in future work.